\documentclass[11pt,a4paper]{article}
\usepackage[hyperref]{acl2019}
\usepackage{times}
\usepackage{latexsym}

\usepackage{url}
\usepackage{amsmath}

\usepackage{graphicx}
\usepackage{multirow}
\usepackage{color}

\aclfinalcopy 

\title{Learning Efficient Lexically-Constrained Neural Machine Translation with External Memory}

\author{Ya~Li${^1}$, Xinyu Liu${^2}$,  Dan Liu$^{1}$, Xueqiang Zhang$^{1}$, Junhua Liu$^{1}$ \\
	\{yali, danliu, xqzhang6, jhliu, siwei, gphu\}@iflytek.com \\
	liuxy.zoe2016@outlook.com \\
	$^{1}$ iFLYTECK Research, Hefei, China \\
	$^{2}$ International Department, The Afflicated School of SCNU, Guangzhou, China.
}

\date{}

\hypersetup{draft}
\begin{document}
\maketitle
\begin{abstract}
Recent years has witnessed dramatic progress of neural machine translation (NMT), however, the method of manually guiding the translation procedure remains to be better explored. Previous works proposed to handle such problem through lexcially-constrained beam search in the decoding phase. Unfortunately, these lexically-constrained beam search methods suffer two fatal disadvantages: high computational complexity and hard beam search which generates unexpected translations. In this paper, we propose to learn the ability of lexically-constrained translation with external memory, which can overcome the above mentioned disadvantages. For the training process, automatically extracted phrase pairs are extracted from alignment and sentence parsing, then further be encoded into an external memory. This memory is then used to provide lexically-constrained information for training through a memory-attention machanism. Various experiments are conducted on WMT Chinese to English and English to German tasks.  All the results can demonstrate the effectiveness of our method. 
\end{abstract}

\section{Introduction}
Neural machine translation is one of the most popular natural language processing tasks over the years, which aims to reduce the difficulty of communication between people from various contries. Recent works \citep{GNMT} have given this goal more realistic meaning through providing human-comparable translation performance on certain domains. Among these excellent translation technics, attention mechanism is verified to be the key for improving the quality of translation \citep{bahdanau2014neural,vaswani2017attention}. \par

Although NMT has made great progress, the effective and friendly interactions between human users and NMT systems still remains a problem, which will be mainly discussed in this paper. For example, users may provide specific words or phrases that are more preferable in the translation results. Another real-world scenario is that some terminologies have totally different translation results in different situations, such as \textit{bank}.   Recent works \citep{hokamp2017lexically,post2018fast,anderson2016guided} regard these problems as lexically-constrianed translation problems and proposed to handle them using lexically-constrained beam search. Unfortunately, existing methods suffer two fatal disadvantages. First, the computational complexities are either exponential \citep{gehring2016convolutional} or linear to the amount of lexical constraints \citep{hokamp2017lexically}. An approach of $\mathcal{O}(1)$ computational complexity in the number constraints has been introduced in a recent work \citep{post2018fast}. However, a sufficiently large beam search size is needed for decoding with large number of lexical constraints. Second, lexically-constrained beam search is a hard method which may cause problems, such as unexpected translation results with right constraints but leaving the other parts inappropriate. \par

In this paper, we propose a novel method to overcome the above mentioned defects of existing methods. Instead of decoding with hard constrained beam search, our model is trained to learn the ability  of lexically-constrained translation with predefined constraints in an external memory. The translation model is then guided to attend to corresponding contents in the external memory with a memory-attention mechanism. For accurate memory-attention, we propose a softmax loss to force the model to attend to the right slots in the memory. Consider the lack of constraints in public training datasets and the easy implementation of our method, an algorithm of automatically extracting constraints is introduced specifically. Compared with previous works, our proposed method mainly has two contributions:
\begin{itemize}
	\item[-] The decoding of our model is as efficient as standard beam search, which can handle large number of lexical constraints with only a normal beam size.
	\item[-] Our lexically-constrained translation is more flexible, which can avoid unexpected translation results even if improper constraints are provided or results with right constraints but leaving the other parts unacceptable.
\end{itemize}

We organize the rest of the paper as follows. Section 2 briefly introduces the background of neural machine translation and lexically-constrained beam search. In section 3, the architecture of training our lexically-constrained model is given and the whole process is detailedly explained. And various empirical experiments are conducted in section 4. Section 5 discusses related works and section 6 concludes the paper.

\section{Lexically-Constrained Beam Search}

\begin{figure*}[t]\label{architecture}
	\begin{center}
		\includegraphics[width=0.8\linewidth]{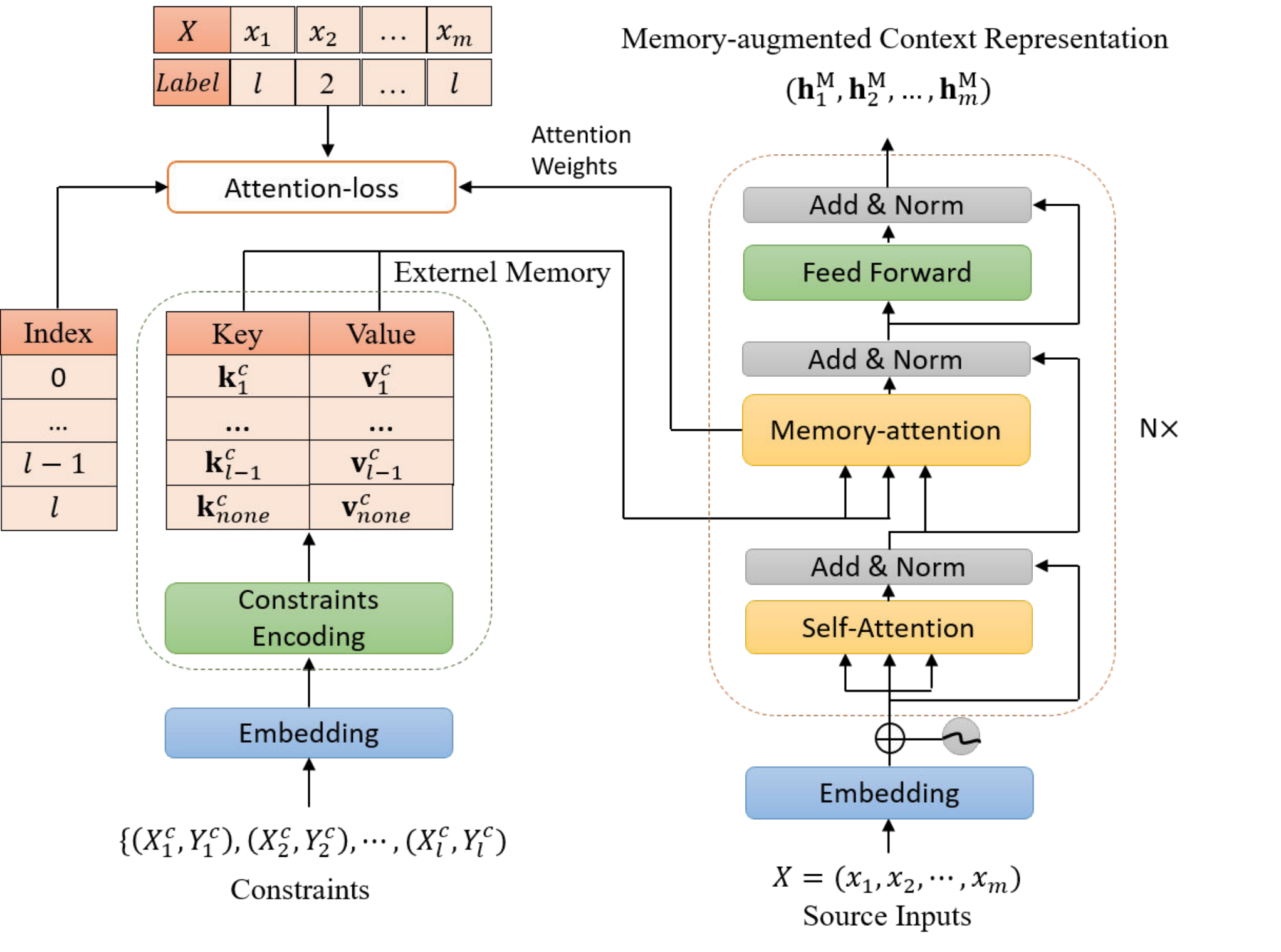}
		\caption{The architecture of our memory-augmented transformer encoder. The constraint pairs are first encoded into an external memory with respect to keys and values. Then the encoder learns to integrate the corresponding contents in the memory with the source inputs through our memory-attention machanism. }
	\end{center}
\end{figure*}

Recent NMT models are mainly based on encoder-decoder neural network architectures, such as RNN-based translation model \citep{bahdanau2014neural} and self-attention based transformer model \citep{vaswani2017attention}. Both of them have dramatically accelerated the development of neural machine translation. \par

For notation simplicity, bold lower case denotes for vectors, capital letters represent sentence sequences. And lower case denotes the individual token in a sequence. Let $(X,Y)$ represents a sentence pair. $X=(x_1,x_2,\cdots, x_m)$ denotes the source sentence of length $m$ and $Y=(y_1,y_2,\cdots,y_n)$ denotes the corresponding target sentence of length $n$. The encoder of the NMT aims to encode the source sentences into vectors of context representations as follows:

\begin{equation}
\mathbf{h_1},\mathbf{h_2},\cdots,\mathbf{h_m} = f_e(x_1,x_2,\cdots,x_m),
\end{equation}
where $\mathbf{h_1},\mathbf{h_2},\cdots,\mathbf{h_m}$ are the corresponding context representations of source sentence $X=(x_1,x_2,\cdots, x_m)$ and $f_e$ is the encoder function. With the context representations, the decoder learns to translate the next target word $y_i$ with previous generated results $y_{<i}=(y_1,y_2,\cdots,y_{i-1})$ as the following:
\begin{equation}
P(y_i) = f_d(y_{<i}, f_a(\mathbf{h_1},\mathbf{h_2},\cdots,\mathbf{h_m})),
\end{equation}
where $f_d$ denotes the decoder function and $f_a$ represents the attention mechanism. With the above equations, the generated probability of target sequence $Y$ can be formulated as follows:
\begin{equation}
P(Y|X) = \prod_{i=1}^{n}f_d(y_{<i}, f_a(\mathbf{h_1},\mathbf{h_2},\cdots,\mathbf{h_m})).
\end{equation}
The training goal of NMT is to maximize the probility $P(Y|X)$ with respect to functions $(f_e,f_d,f_a)$.

In the inference phase, it is time consuming and requires a large memory space if we want to find the best result from the whole search space. If the vocabulary size is $v$, the size of the search space is $v^n$, where $n$ is the length of the target. Fortunately, beam search is proposed to efficiently approximate the best result using a heuristic search algorithm, which selects the top $k$ best beams every inference time. Note that $k$ is regarded as the beam size. \par

Lexically-constrained beam search is a variant of normal beam search which can generate results containing predefined constraints. However, this may sacrifice the efficiency and accuracy of decoding. For example, the decoding complexity of Grid Beam Search (GBS) \citep{hokamp2017lexically} is linear to the number of constraints with effective beam size of $k\times(C+1)$. Alghough the dynamic beam allocation (DBA) \citep{post2018fast} has a similar time complexity to the normal beam search, it requires a large beam size which is indirectly propotional to the number of constraints. Additionally, the lexcially-constrained beam search forces the decoding results to contain constraints, which dramatically limit the search space of beam search especially with long phrase constraints. Consequently, the performance of the final generated results is degraded.

\section{Lexically-Constrained Neural Machine Translation with External Memory}

\begin{figure*}[t]\label{phrase_extract}
	\begin{center}
		\includegraphics[width=1\linewidth]{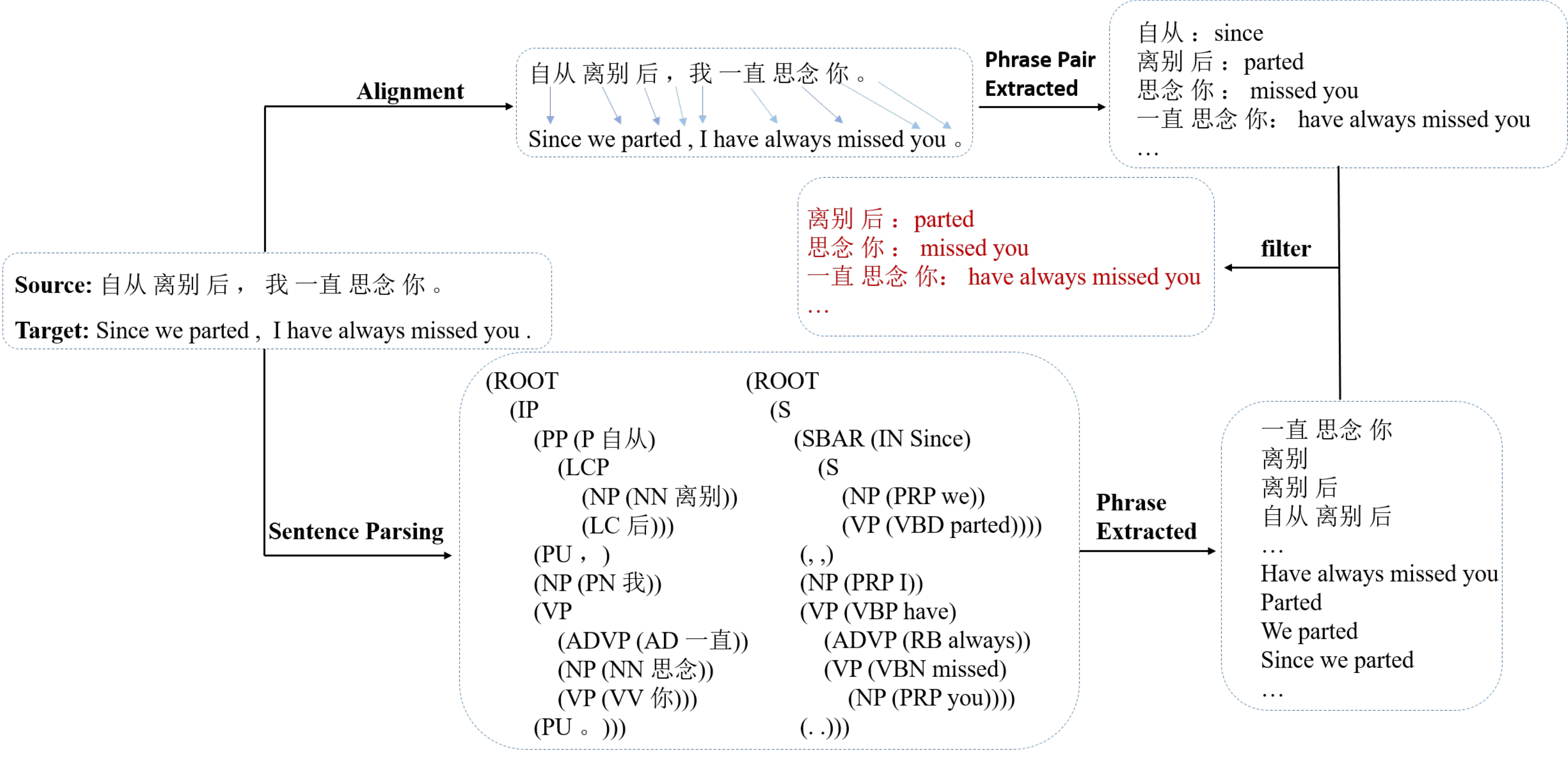}
		\caption{The pipeline of extracting phrase pairs. }
	\end{center}
\end{figure*}

In this section, we discuss the details of our lexically-constrained neural machine translation with external memory. Instead of forcing the beam search to generate results with constraints in the inference time, we propose to learn such an ability of translating with lexical constraints. Let $M$ be the external memory of constraint pairs $\{(X^c_1,Y^c_1),(X^c_2,Y^c_2),\cdots,(X^c_l,Y^c_l)\}$. Note that $(X^c_i, Y^c_i)$ is the $i$-th lexically-constrained pair with source phrase $X^c_i$ to be translated to target phrase $Y^c_i$ and $l$ represents the number of constraints. One constraint $X^c_i$ or $Y_i^c$ may include several words. The intuition behind our algorithm is that the constraints are expected to be integrated into the context representations in the encoder, such that

\begin{equation}
\mathbf{h^M_1},\mathbf{h^M_2},\cdots,\mathbf{h^M_m} = f_e(x_1,x_2,\cdots,x_m, f_M(M)),
\end{equation}
where $f_M$ is the attention mechanism to the memory of the constraints and $\mathbf{h^M_1},\mathbf{h^M_2},\cdots,\mathbf{h^M_m}$ is the memory-augmented context representations. 

When decoding, the constrained translated results are generated using the information from the memory-augmented context representations. The training goal is to maximize the following probability with respect to functions $(f_a,f_d,f_e,f_M)$:

\begin{equation}
 P(Y|X,M) = \prod_{i=1}^{n}f_d(y_{<i}, f_a(\mathbf{h^M_1},\mathbf{h^M_2},\cdots,\mathbf{h^M_m})).
\end{equation}

\subsection{Architecture}
In this paper, the architecture of our lexically-constrained neural machine translation is based on Transformer \citep{vaswani2017attention}, which is one of the most popular NMT models. However, our proposed algorithm can also be implemented with RNN-based translation model. \par

The detailed architecture is shown in Figure 1. Comparing to the standard transformer architecture, we add one external memory block with an attention-loss and one memory-attention layer in the encoder. The other parts and the decoder of our model is the same as those of standard transformer. The external memory block aims to encode each constraint pair $(X_i^c,Y_i^c)$ into a key and value pair $(\mathbf{k}^c_i, \mathbf{v}_i^c)$. The encoding of constraint pairs can be simply implemented using the average of word embeddings in the phrase, such that
\begin{equation}
\begin{aligned}
\mathbf{k}_i^c &= \frac{1}{|X_i^c|}\sum_{x^c_i \in X_i^c}f_s^{emb}(x_i^c), \\
\mathbf{v}_i^c &= \frac{1}{|Y_i^c|}\sum_{y^c_i \in Y_i^c}f_t^{emb}(y_i^c), 
\end{aligned}
\end{equation} 
where $f_s^{emb}$ and $f_t^{emb}$ denote the embedding function of source language and target language. $|X_i^c|$ refers to the length of the source constraint $X_i^c$ and $x^c_i \in X_i^c$ means $x_i^c$ is one token in source constraint $X_i^c$. Note that the source constaints and the source inputs share the same embedding function, and so does the target constraints and the target inputs. Indeed the encoding of the constraints can also be implemented with LSTM or multi-layers of self-attention, however, the average of word embeddings can work well enough from our empirical results. Additionally, using average of word embeddings increases less training parameters and has a higher training efficiency than LSTM or multi-layers of self-attention.     \par

\subsection{Automatically Extracted Constraints} \label{extract_constraints}
As shown in the architecture, the training of our lexically-constrained NMT needs corresponding constraints for input and ouput sequences. Unfortunately, existing training datasets of NMT hardly have these constraints because of large cost of time. Consequently, we propose an automatical method for extracting constraint pairs for any bilingual datasets. \par

Figure 2 shows an example of extracting phrase pairs from bilingual training data and the extracted phrase pairs are then used as constraints for training the lexically-constrained NMT. The process of extracting phrases mainly includes two parts: alignment and parsing. The sentence alignment is used to extract possible phrase pairs. If several alignments from source to target are consistent in positions, the combination of these alignments is viewed as one phrase pair.
For example, the alignments (sinian $\rightarrow$ missed, ni $\rightarrow$ you) should be combined into one phrase pair (sinian ni, missed you). However, some extracted phrase pairs are noisy, which are misaligned or not meaningful. To avoid such phrase pairs, we propose to filter the phrase pairs extracted from alignment through sentence parsing. Sentence parsing can provide more canonical phrases, such as noun phrase, verb phrase etc. If the phrases extracted from alignment also exists in the candidate phrases from corresponding parsing, these phrases are reserved as final constraints for the current sentence pair. 

\subsection{Memory-Attention Mechanism}
The added memory-attention layer in the encoder block is one multi-head attention layer \citep{vaswani2017attention} with encodings of the inputs as queries and $(\mathbf{k}_i^c,\mathbf{v}_i^c)$ from memory as keys and values. 

For training, the constraints of the sentences from one batch are simultaneously encoded into the external memory. Therefore, the memory-attention mechanism needs to find the most correlated constraint with respect to the inputs among all constraints in the batch. Suppose there are total $l$ constraints including $(l-1)$ normal constraints $\{(\mathbf{k}_i^c, \mathbf{v}_i^c),i=1,2,\cdots,l-1\}$ and one special constraint $(\mathbf{k}_{none}^c, \mathbf{v}_{none}^c)$ in the $l$-th slot of the memory. This special constraint provides no information and is used for input tokens which have no related constraints in the memory. Let $\mathbf{K}=[\mathbf{k}_1^c, \mathbf{k}_2^c, \cdots, \mathbf{k}_{l-1}^c, \mathbf{k}_{none}^c] \in \mathcal{R}^{d_k \times l}$ be a matrix of all keys with each vector $\mathbf{k}_i^c$ as the column. Additionally, matrix $\mathbf{V}=[\mathbf{v}_1^c,\mathbf{v}_1^c,\cdots,\mathbf{v}_{l-1}^c, \mathbf{v}_{none}^c] \in \mathcal{R}^{d_k \times l}$ represents all values. Note that $d_k$ represents the dimention of keys and values. For any query $\mathbf{q}_j$ of the input token $x_j$, the memory-attention can be formulated as one scaled dot-product attention. 
\begin{equation}
\begin{aligned}
& \mathbf{p}_j = softmax(\frac{\mathbf{q}_j^\top \mathbf{K}}{\sqrt{d_k}}) \\
&Attention(\mathbf{q}_j, \mathbf{K}, \mathbf{V}) = \mathbf{p}_j \mathbf{V}^\top.
\end{aligned}
\end{equation}

Note that $\mathbf{p}_j$ is a $l$-dimensional probability vector $\mathbf{p}_j = [p_{j1}, p_{j2}, \cdots, p_{jl}]$, which represents the attention weights between $\mathbf{q}_j$ and $\mathbf{K}$. \par

The goal of the memory-attention mechanism is to attend to the most correlated slot in the memory and integrates the context information of the target constraints into the corresponding queries. If $l$ is a large number, the memory can provide much information about the translated results. Consequently, the model can be easily learned to be over-reliant on the memory and the performance will decrease without any constraints provided comparing to standard transformer. Conversely, the model will ignore the information in the memory if $l$ is too small. To avoid such situation, we propose an additional attention loss for the memory-attention mechanism as illustrated in Figure 1. \par 

For each token $x_j$ in the inputs, we generate a label $s_j$ for memory-attention. This label indicates the index of the slot in the memory, to which the token should attend. Note that if one token has no corresponding slot in the memory, it will attend to the special slot $(\mathbf{k}_{none}, \mathbf{v}_{none})$. Consequently, the attention label of this token is $l$. The additional attention loss for token $x_j$ can be formulated as follows:
\begin{equation}\label{attention_loss}
Att\_loss = \frac{1}{N} \sum_{j=1}^N -log(p_{js_j}),
\end{equation}
$N$ is the number of tokens in the batch and $p_{js_j}$ refers to the attention probability of the $s_j$-th slot for token $j$. This loss is then minimized together with the main loss in the output of the decoder. With the proposed memory attention loss, the attention ability can be learned more efficiently and accurately even if with just a small amount of constraints.

\begin{table}
	\begin{center}
		\label{block_select}
		\begin{tabular}{|c|c|c|}
			\hline
			Type & $\text{newstest}\_\text{zhen}$ & $\text{newstest}\_\text{ende}$  \\
			\hline
			Sentence & $2001 $ & $3003 $   \\
			\hline
			Phrase & $5434 $ & $6806 $  \\
			\hline
			Word & $12881 $ & $16173 $  \\
			\hline
			Sub$\_$word & $14606 $ & $20890$ \\
			\hline
		\end{tabular}
		\caption{Statistics for $\text{newstest}\_\text{zhen}$ and  $\text{newstest}\_\text{ende}$ with respect to the number of sentences, number of phrases, number of words in all phrases and number of sub$\_$words in all phrases from the target language.}
	\end{center}
\end{table}

\begin{table}
	\begin{center}
		\label{block_select}
		\begin{tabular}{|c|c|c|}
			\hline
			Block ID & Zh $\rightarrow$ En & En $\rightarrow$ De  \\
			\hline
			1 & $28.3 $ & $31.5 $   \\
			\hline
			2 & $28.6 $ & $31.8 $  \\
			\hline
			3 & $28.4 $ & $31.5 $  \\
			\hline
			4 & $28.3 $ & $31.4$ \\
			\hline
			5 & $28.2 $ & $31.5$ \\
			\hline
			6 & $28.3 $ & $31.3$ \\
			\hline
		\end{tabular}
		\caption{BLEU scores comparison between models with different encoder blocks added by a memory-attention layer.}
	\end{center}
\end{table}

\begin{table}
	\begin{center}
		\label{block_select}
		\begin{tabular}{|c|c|c|c|c|c|}
			\hline
			\multicolumn{2}{|c|}{Task} & \multicolumn{3}{|c|}{$\text{newstest}\_\text{ende}$} \\
			\hline
			\multicolumn{2}{|c|}{Method} & Base & DBA & LCNMT\\
			\hline
			\multirow{3}*{Ratio} & $30\%$ & $27.3 $  & $28.0$ & $\mathbf{30.2}$ \\
			\cline{2-5}
			& $50\%$ & $27.3 $  & $29.6$ & $\mathbf{31.8}$ \\
			\cline{2-5}
			& $70\%$ & $27.3$  & $32.2$ & $\mathbf{33.8}$ \\
			\hline
		\end{tabular}
		\caption{BLEU scores comparison of different methods on $\text{newstest}\_\text{ende}$. Three different raitos $30\%$, $50\%$, $70\%$ of automatically-extracted phrases are randomly selected as constraints.}
	\end{center}
\end{table}

\begin{table}
	\begin{center}
		\label{block_select}
		\begin{tabular}{|c|c|c|c|c|}
			\hline
			\multicolumn{2}{|c|}{Task} & \multicolumn{3}{|c|}{$\text{newstest}\_\text{zhen}$} \\
			\hline
			\multicolumn{2}{|c|}{Method} & Base & DBA & LCNMT\\
			\hline
			\multirow{3}*{Ratio} & $30\%$ & $24.2 $  & $\mathbf{27.3}$ & $27.0$ \\
			\cline{2-5}
			& $50\%$ & $24.2 $  & $28.4$ & $\mathbf{28.6}$ \\
			\cline{2-5}
			& $70\%$ & $24.2$  & $29.6$ & $\mathbf{29.9}$ \\
			\hline
		\end{tabular}
		\caption{BLEU scores comparison of different methods on $\text{newstest}\_\text{zhen}$. Three different raitos $30\%$, $50\%$, $70\%$ of automatically-extracted phrases are randomly selected as constraints.}
	\end{center}
\end{table}

\section{Experiments}
This section shows various experimental results to demonstrate the effectiveness of our proposed lexically-constrained neural machine translation (LCNMT). Two translation directions, Chinese $\rightarrow$ English and English $\rightarrow$ German, are trained on the WMT'17 corpora \citep{bojar2017findings}. The corpora is first processed by filtering noisy bilingual sentences, such as sentence pairs with abnormal length ratio,  sentences pairs with target language the same as the source language, and sentence pairs appear in the corpora multiple times. For English and German corpora, we tokenize them with Moses tokenizer \footnote{http://www.statmt.org/moses/}. And the Chinese corpora is tokenized with LTP tokenizer \footnote{https://github.com/HIT-SCIR/ltp}. All tokenized corpora is then processed with sub-word  \cite{sennrich2015neural} using 40k merge operations. We implement our algorithm based on Transformer from Tensor2Tensor \footnote{https://github.com/tensorflow/tensor2tensor}. All experiments are run on 4 GPUs using a base model with batch size of 9000 tokens and the BLEU scores are evaluated on detokenized results using SAREBLEU \cite{post2018call}. As the efficiency of GBS \citep{hokamp2017lexically} is low and DBA \citep{post2018fast} has a comparable performance to GBS, therefore,
our LCNMT is compared with two methods, base model of transformer \citep{vaswani2017attention} and DBA \citep{post2018fast}. \par


\subsection{Setup of Memory-Attention}
To keep the training efficiency close to that of the standard transformer, just one of the encoder block is required to have a memory-attention layer. To confirm the best encoder block to attend to the memory, we first train models with different encoder blocks added by a memory-attention layer. \par

We test our Chinese $\rightarrow$ English models on newstest2017 and our English $\rightarrow$ German models on newstest2014. For notation simplicity, we denote newstest2017 for Chinese $\rightarrow$ English as $\text{newstest}\_\text{zhen}$ and newstest2014 for English $\rightarrow$ German as $\text{newstest}\_\text{ende}$.  Table 1 shows some statistics of $\text{newstest}\_\text{zhen}$ and $\text{newstest}\_\text{ende}$ with respect to the number of sentences, number of extracted phrases, number of words in all phrases and number of sub$\_$words in all phrases from the target language. The phrases are extracted as discussed in section \ref{extract_constraints}. From the statistics, the average numbers of sub$\_$word constraints for $\text{newstest}\_\text{zhen}$ and $\text{newstest}\_\text{ende}$ per sentence are computed as $7.30$ and $6.96$. Note that some sentences may have no phrases because of no accurate alignments found with alignment and parising. \par

We randomly select $50\%$ of the extracted phrases as constraints and compare the performance of the models with different blocks added by a memory-attention layer. The experimental results are shown in Table 2. From the results, we can conclude that adding the memory-attention layer to the second block of the encoder is the best choice.   \par


\begin{table*}
	\begin{center}
		\label{example}
		\begin{tabular}{l}
			\hline
			\textbf{Source:} We don’t have to rush into surgery that is irreversible. \\
			\hline
			\textbf{Constraints:} sich nicht mehr rückgängig machen lässt. \\
			\hline
			\textbf{Reference:} Und man muss nicht übereilt eine Operation vornehmen, die sich nicht mehr rückgängig \\
			machen lässt. \\
			\hline
			\textbf{Base:} Wir müssen uns nicht in eine Operation stürzen, die unumkehrbar ist. \\
			\hline
			\textbf{DBA:} Wir müssen nicht in eine irreversible Operation voreilig greifen. sich nicht mehr rückgängig \\
			machen läss. \& \# 160; \& \# 160; \& \# 160; \& \# 160; \\
			\hline
			\textbf{LCNMT:} Wir müssen uns nicht in eine Operation stürzen,   die sich nicht mehr rückgängig machen \\
			lässt. \\
			\hline
		\end{tabular}
		\caption{An example of results comparison between different methods for task English $\rightarrow$ German.}
	\end{center}
\end{table*}

\subsection{Performance on WMT}
According to the above analysis, the memory-attention layer will be added to the second encoder block. We evaluate all methods on both $\text{newstest}\_\text{zhen}$ and  $\text{newstest}\_\text{ende}$. For comparison of the performance with various number of constraints, we randomly
select different ratios $30\%,50\%,70\%$ of the phrases as constraints for decoding. For all methods, beam size 12 is used for decoding which is sufficient for beam search. \par

Table 3 and Table 4 shows the performance comparison of all methods in terms of BLEU scores on  $\text{newstest}\_\text{ende}$ and $\text{newstest}\_\text{zhen}$ respectively. For task English $\rightarrow$ German,  we can conclude that all lexically-constrained methods can outperform baseline method. Additionally, our proposed LCNMT performs the best with different ratios of constraints.  For task Chinese $\rightarrow$ English,  our LCNMT has a comparable performance with DBA.  Our LCNMT is a soft lexically-constrained translation method which can generate constrained results with a large probability and guarantee the fluency of the results simutaneously. An example is given in Table 5. From the results, we find that all methods can generate results with given constraints except the baseline method. However, the generated results of DBA are less fluent which contains some unexpected generations. 



\section{Related Work}
The establishment of one efficient and effective machine translation system is attractive over the decades. Although systems based on statistical machine translation \citep{callison2005introduction} have been used in real life, the unpromising performance makes it difficult to be promoted. Recent works \citep{cho2014properties,gehring2016convolutional,bahdanau2014neural,vaswani2017attention,lample2017unsupervised,lample2018phrase} of neural machine translation have made this possible. \citep{bahdanau2014neural} proposed an attention mechanism for encoder-decoder neural machine translation system, which can sufficiently explore the context representation in the source sentences. Transformer \cite{vaswani2017attention} is a more promising neural machine translation architecture with self-attention, which can achieve faster training speed and better performance. \par 

Several works \citep{anderson2016guided,hokamp2017lexically,post2018fast} have discussed lexically-constrained beam search from different aspects. \citep{anderson2016guided} applies constrained beam search to image caption tasks, which aims to handle out-of-domain scenes or objects. An finite-state machine with the states representing the completed constraints is cooperated with beam search. However, the decoding complexity is exponential to the number of constraints. An improved Grid Beam Search method is proposed in \citep{hokamp2017lexically}, which extends an individual beam to the size of number of constraints for exhaustively searching results with completed constraints. The time complexity of GBS is linear to the number of constraints. Additionally, parallel implementation of GBS is troublesome because of variant beam size caused by different number of constraints for each sentence. \citep{post2018fast} makes a significant improvement over GBS by dynamic beam allocation, which can reduce the time complexity to $\mathcal{O}(1)$ in the number of constraints. Unfortunately, the beam size is required to be much larger than the number of constraints and this hard beam search can sometimes generate strange results. \par 

External memory has been used in several works \cite{zhao2018phrase,meng2018neural,feng2017memory} to enhance the quality of neural machine translation. For example, \citep{zhao2018phrase} proposes to extract phrase table as recommendation memory for neural machine translation. However, this kind of phrase table is too noisy, which is also mentioned in \citep{post2018fast}. \citep{feng2017memory} proposes to store the hidden context information into the memory, which can be used to calculate an additional probability of target word. Both of these two methods require a high quality translation alignment. \cite{pham2018towards} proposes to annotate the source sentences with experts and use a copy-generator for rare word translation. However, the strong copy ability may cuase the loss of fluency. And \citep{meng2018neural} aims to improve the performance of NMT by maintaining a updatable memory.

\section{Conclusions}
In this paper, we propose an algorithm of training lexically-constrained translation with external memory. Compared with DBA, our method can decode more efficiently with a soft lexically-constrained memory. For better implementation of our method, we propose a procedure for automatically extracting phrases which can provide constraints for any bilingual corpus. An memory-attention loss is ultilized to force accurate memory-attention with a small amount of constraints. Experimental results can demonstrate the effectiveness of our LCNMT. 


\bibliography{acl2018}
\bibliographystyle{acl_natbib}

\end{document}